\documentclass[11pt,a4paper]{article}
\usepackage{authblk}
\usepackage[hyphenbreaks]{breakurl}
\usepackage[hyperref]{emnlp2020}
\usepackage{times}
\usepackage{latexsym}
\usepackage{clrscode}
\usepackage{arydshln}
\usepackage{graphicx}  
\usepackage{bbm}
\usepackage{siunitx}
\usepackage{pslatex}
\usepackage{booktabs}
\usepackage{multirow}
\usepackage{subcaption,amsfonts,dcolumn}
\usepackage{amsmath}
\usepackage{hyperref}
\usepackage{comment}
\usepackage{caption}

\newcommand{\method}{\proc{cod3s}}
\newcommand{\acronym}{\proc{cod3s}:
\textbf{CO}nstrained \textbf{D}ecoding with \textbf{S}emantic \textbf{S}entence \textbf{S}ignatures}

\newcommand\scalemath[2]{\scalebox{#1}{\mbox{\ensuremath{\displaystyle #2}}}}
\usepackage[hang,flushmargin]{footmisc}
\usepackage{microtype}

\aclfinalcopy 
\def\aclpaperid{3457} 

\usepackage{titlesec}
\usepackage{amssymb,amsmath,amsthm,enumitem}

\titlespacing\section{0pt}{12pt plus 2pt minus 0pt}{4pt plus 2pt minus 0pt}
\newcommand{\mysubsection}{\paragraph}

\newcommand{\appropto}{\mathrel{\vcenter{
  \offinterlineskip\halign{\hfil$##$\cr
    \propto\cr\noalign{\kern2pt}\sim\cr\noalign{\kern-2pt}}}}}
\title{\method{}: Diverse Generation with Discrete Semantic Signatures}

\author[1]{\textbf{Nathaniel Weir}}
\author[2]{\textbf{Jo\~{a}o Sedoc}}
\author[1]{\textbf{Benjamin Van Durme}}
\affil[1]{Department of Computer Science, Johns Hopkins University}
\affil[2]{Department of Technology, Operations, and Statistics, New York University}
\affil[ ]{\texttt{\{nweir, vandurme\}@jhu.edu, jsedoc@stern.nyu.edu}}

\date{}

\begin{document}
\maketitle
\begin{abstract}
We present \method, a novel method for generating semantically diverse sentences using neural sequence-to-sequence (seq2seq) models.
Conditioned on an input, seq2seq models typically 
produce
semantically and syntactically homogeneous sets of sentences and thus perform poorly on one-to-many sequence generation tasks. Our two-stage approach
improves output diversity by conditioning generation on locality-sensitive hash (LSH)-based \textit{semantic sentence codes} whose Hamming distances highly correlate with human judgments of semantic textual similarity. 
Though it is generally applicable, we apply \method{} to causal generation,
the
task of 
predicting a proposition's plausible causes or effects. We demonstrate through automatic and human evaluation that responses produced using our method exhibit improved diversity without degrading task performance.
\end{abstract}

\section{Introduction}

Open-ended sequence generation problems 
such as dialog, story generation, image captioning, 
or causal generation 
pose a practical challenge to neural sequence-to-sequence (seq2seq) models, as they necessitate a diverse set of predicted outputs.
The typical sampling method for seq2seq decoding is beam search, which produces a set of candidate sequences that generally have high syntactic, lexical, and semantic overlap. 

\begin{figure}[!t]
    \centering
    \includegraphics[width=\columnwidth]{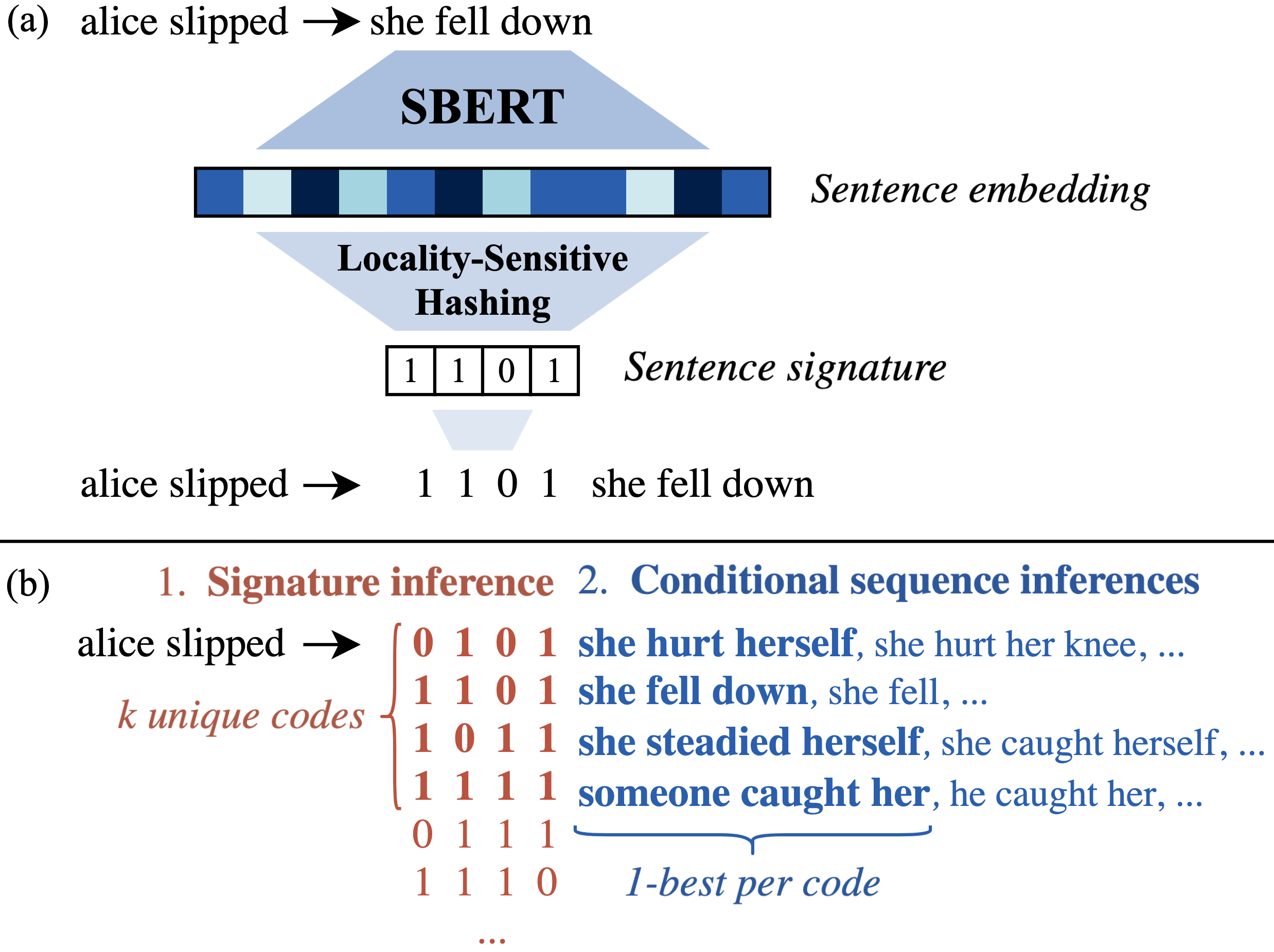}
    \caption{Overview of the \method{} method. In training \textbf{(a)}, the target side is prefixed with a discrete signature computed using locality-sensitive hashing (LSH) of the target's SBERT
    embedding. At inference \textbf{(b)}, a beam search is conditioned on each of $k$ decoded signatures. }
    \label{fig:overview}
    \vspace{-0.5em}
\end{figure}
Recent methods for improved diversity generation make slight modifications to the neural architecture or beam search algorithm~\cite{Xu2018DPAGEDP,li2016simple}, or impose lexical constraints during decoding~\cite{post-vilar-2018-fast,hu-etal-2019-improved}. \citet{shu-etal-2019-generating} propose the use of \textit{sentence codes}, a technique in which generation is conditioned on a discrete code that aims to induce diversity in syntax or semantics. While their approach is effective for syntactic codes, it is less so for semantics.

In this work, we introduce an improved method for diverse generation conditioned on inferred sentence codes that explicitly capture meaningful semantic differences.
We use the contextual sentence embeddings from Sentence-BERT~\cite[SBERT;][]{reimers-2019-sentence-bert}, the cosine distances between which correlate highly with human scalar judgments of semantic textual similarity (STS). 
We construct discrete codes from these embeddings using \textit{locality-sensitive hashing}~\cite{indyk1998approximate,charikar2002similarity}, producing short binary signatures whose Hamming distances well-preserves the cosine distances between
inputs. 

Our method induces a bitwise hierarchy of semantic bins whose similarities in signature imply similarities in semantics.
Conditioning generation on a signature as a target-side prefix indicates the bin into which the generated sequence falls. We implement a two-stage decoding process that (1) infers the most relevant signatures and (2) decodes sequences 
via separate prefix-conditioned beams. We term our method \acronym. 

We demonstrate the effectiveness of \method{} in the context of causal sequence generation~\cite{causalbank} through BLEU- and cosine-based diversity measures as well as human evaluation.

\section{Related Work}
We draw inspiration from recent work in multilingual machine translation (MT)~\cite{ha2016toward} and domain adaptation~\cite{chu2019multilingual} in which a language code 
(e.g. \textbf{en}, \textbf{de}) is prepended to the target to guide generation.
Our method for encoding sentence diversity
is closely related to MT work by \citet{shu-etal-2019-generating}, who condition generation on prefixed \textit{sentence} codes.
They improve the syntactic diversity of sampled translations using codes produced from improved semantic hashing~\cite{kaiser2018discrete} with a TreeLSTM-based autoencoder. 
Their experiments with semantic coding via clustering of BERT~\cite{devlin-etal-2019-bert} and FastText~\cite{bojanowski-etal-2017-enriching} embeddings lead to negligible or negative effects.
Outside of MT, \citet{keskar2019ctrl} in a similar vein condition on manually categorized ``control codes" that specify style and content, and \citet{mallinson2019controllable} condition on annotated syntactic or lexical change markers that can be learnt from data.
We refer readers to \citet{ippolito-etal-2019-comparison} for an overview of diverse decoding methods. Few to our knowledge explicitly and effectively encode open-domain semantic diversity.

Text-based causal knowledge acquisition is a well-studied challenge in NLP~\cite{radinsky2012learning}. Recent efforts have investigated \textit{open ended} causal generation using neural models \cite{bosselut2019comet,causalbank}. The latter train a 
conditional generation model to propose cause or effect statements for a given proposition. The model is trained on the co-released corpus CausalBank, which comprises causal statements harvested from English Common Crawl~\cite{buck2014n}.

Applications of LSH~\cite{indyk1998approximate,charikar2002similarity} in NLP began with \citet{ravichandran-etal-2005-randomized} who demonstrated its use in fast lexical similarity comparison; later, \citet{van-durme-lall-2010-online} showed such hashing could be performed online. More similar to our use case, \citet{petrovic-etal-2010-streaming} binned tweets via LSH to enable fast \emph{first story detection}.  Most related to ours is work by \citet{guu-etal-2018-generating}, who describe a generative sentence model that edits a `prototype' sentence using lexically similar ones retrieved via LSH.
\section{\method{} Approach}
Our signature construction method, depicted in \autoref{fig:overview}(a), 
 produces a sequence of bits 
that
 collectively
 imply a highly specific bin of sentences with similar semantic meaning. This is accomplished by encoding sentences into high-dimensional vectors that encode degrees of semantic difference and then discretizing the vectors in a way that approximately preserves the difference.
\mysubsection{Semantic Embedding Model}
We embed a sentence using the contextual encoder Sentence-BERT ~\cite[SBERT;][]{reimers-2019-sentence-bert}, a siamese network trained to produce 
embeddings whose cosine similarity approximates the semantic textual similarity (STS) of the underlying sentences.
We select this \textit{single sentence} encoder over other popular encoders, e.g. BERT,
which
best encode concatenations of pairs of sentences and therefore do not produce individual embeddings that encode semantic difference retrievable under vector similarity metrics~\cite{reimers-2019-sentence-bert,shu-etal-2019-generating}.
The cosine similarity of embeddings from SRoBERTa-L, the instance of SBERT that we use as our \method{} encoder, has a Spearman $\rho$ correlation of $.863$ with human STS judgements from STSbenchmark~\cite{cer-etal-2017-semeval}.\footnote{We use the \href{https://github.com/UKPLab/sentence-transformers}{released} SRoBERTa instance that was fine-tuned on natural language inference (NLI) and then STS.
}
 We provide a list of cosine/STS correlations using other models in \autoref{app:bins-appendix}.\footnote{We refer readers to  \citet{reimers-2019-sentence-bert} (Sec.4) for a comprehensive overview using other STS datasets.}

\mysubsection{Discretization via LSH}
Locality-sensitive hashing~\cite[LSH;][]{indyk1998approximate} maps high-dimensional vectors into low-dimensional sketches for quick and accurate similarity comparison under measures such as cosine or Euclidean distance. We use the  popular variant by \citet{charikar2002similarity}, which computes a discrete $b$-bit signature 
$\texttt{LSH}(\vec{v}) = [\texttt{LSH}_1(\vec{v}), \dots \texttt{LSH}_b(\vec{v})]$.
\autoref{app:LSH} provides an overview of this approach. The Hamming distance between two LSH signatures approximates the cosine distance of the underyling vectors: 
\[
\scalemath{.9}{
\text{cos}(\vec{u},\vec{v}) = \dfrac{\vec{u}\cdot \vec{v}}{|\vec{u}| |\vec{v}|} \approx \text{cos}\biggl(  \frac{\pi}{b}\sum_{i=1}^b \mathbbm{1}\{\texttt{LSH}_i(\vec{u}) \neq \texttt{LSH}_i(\vec{v})\}\biggr)
}
\]
This approximation degrades with coarser-grained signatures, as shown by the drop in STS correlation in \autoref{tab:full-sts} (right columns) for LSH with fewer bits. 

\paragraph{A Hierarchy of Signatures}
Using LSH 
on
SBERT embeddings whose cosine similarity correlates highly with STS induces a \textit{hierarchy of semantic bins}; the $i+1$th bit partitions each of a set of $i$-bit bins in two. Bins whose signatures differ by few bits have higher semantic overlap, and as the bitwise distance between two signatures increases, so does the difference in meaning of the underlying sentences.  Sentences that hash to the same bin---particularly for longer signatures---have very high SBERT cosine similarity and are thus likely semantically homogeneous.  
\begin{table}[!t]
    \small
    \centering
    \begin{tabular}{ccccccc}
    \toprule
    & \textbf{Cosine} & \multicolumn{5}{c}{\textbf{$b$-Bit LSH Hamming Distance}} \\ \cmidrule(lr){2-2}\cmidrule(lr){3-7}
    & 1024D & 256b & 128b & 32b & 16b & 8b \\ \midrule
    \textbf{STS $\rho$} & .863 & .845 & .828 & .742 & .652 & .549\\
    \bottomrule
    \end{tabular}
    \caption{Correlation of SRoBERTa-L embedding cosine distance and LSH Hamming distance with STS judgements from STSBenchmark.
    }
    \label{tab:full-sts}
\end{table}
\paragraph{Diverse Decoding Using Signatures}
Given source and target sentences $x,y$, we compute the $b$-bit signature $s^y = \texttt{LSH}(\texttt{SBERT}(y))$. We then train a model to decode the concatenated sequence $[s^y \ y]$, with the $s^y$ treated as a $b$-length sequence of individual $0/1$ tokens.
At inference time, we decompose the typical conditional decision problem
$ \hat{y} = \mathop{\text{argmax}}_{y} \{\log p(y \mid x)\}$ 
 into two steps:
\vspace{-.15em}
\[
\scalemath{.9}{
    \hat{s} = \mathop{argmax}_{s} \{\log{p(s \mid x)} \}; \  \ 
    \hat{y} = \mathop{argmax}_{y} \{\log{p(y \mid x, \hat{s})}\}
}
\vspace{-0.3em}
\]
As previous work associates the strength of a causal relationship with pointwise mutual information (PMI)~\cite{gordon-etal-2012-semeval}, we modify our objective to maximize the MI between $x$ and each of $s$ and $y$; we adapt the \textbf{MMI-bidi} objective from \citet{li-etal-2016-diversity}:
\begin{align}
\vspace{-0.4em}
    \hat{s} &= \mathop{argmax}_{s} \{\log{p(s \mid x)} + \lambda_s\log{p(x \mid s)} \} \\
    \hat{y} &= \mathop{argmax}_{y} \{\log{p(y \mid x, \hat{s})}+ \lambda_y\log{p(x \mid y)}\}
\vspace{-0.4em}
\end{align}
As shown in \autoref{fig:overview}(b), we first decode the $k$-best distinct sentence codes $\hat{s}_1,\dots\hat{s}_k$ as in Eq. 1.
We then perform $k$ conditional inferences in Eq. 2; we take the 1-best sentence from each to produce $\hat{y}_1,\dots \hat{y}_k$.
For both signature and sentence decoding, we follow \citeauthor{li-etal-2016-diversity} and sample an $n$-best list from the forward score $\log{p(s \mid x)}$ (resp. $\log{p(y \mid x, \hat{s}})$) before re-ranking with the added $\lambda$-weighted backward score.\footnote{We find effective values $\lambda_s = 1000,\lambda_y=0.3$ for 16-bit \method{} using qualitative examination of predictions.} 
We approximate the forward scores 
using length-normalized beam search with beam size 100 for signatures and 40 for sentences.
While $\log{p(s \mid x)}$ and $\log{p(y \mid x,s)}$ can be scored using a single forward model, we find it beneficial to train two, so that the first only learns to score signatures. 

\paragraph{Hamming Distance Threshold}
As sentences 
whose signatures differ by few bits show to have highly similar semantics, we impose a threshold heuristic for decoded signatures $\hat{s}_1,\dots, \hat{s}_k$:
$\min_{i\neq j} {D(\hat{s}_i,\hat{s}_j)} > t$,
where $D(\cdot)$ is Hamming distance.\footnote{We find the threshold $t=2$ best for 16-bit \method{}.} We enforce this using a greedy algorithm that considers higher-scoring signatures first, keeping those that satisfy the threshold given the currently kept set and removing those that violate it. 

Taken as a whole, our decoding approach 
aims to generate the single highest-scoring applicable response that falls in each of the N-best inferred \textit{sufficiently different} semantic bins. The threshold parameter thus provides a way to effectively tune the model to a desired level of semantic diversity.
\section{Experiments}
We apply \method{} to the task of open-ended causal generation for free-form textual inputs as considered by \citet{causalbank}. Given an input statement,
the model must suggest a \textit{diverse} set of possible causes 
or effects.
We train models on sentence pairs from \citeauthor{causalbank}'s released dataset, CausalBank, 
 which is scraped from Common Crawl using templatic causal patterns. Following their work,
 we use 10 million sentence pairs that match the patterns ``X, so Y" to train cause-to-effect models and ``X because Y" for effect-to-cause models.

We experiment with 16-bit LSH signatures of SBERT embeddings.\footnote{Statistics describing the distribution of the 10M training targets into signature bins 
are given in \autoref{app:bins-appendix}.}
After prepending target-side bit signatures, pairs are encoded with byte-pair encoding~\cite[BPE;][]{sennrich-etal-2016-neural} using a vocabulary size of 10K. 
We train Transformer models~\cite{vaswani2017attention}  using the \proc{fairseq} library~\cite{ott2019fairseq}.  \autoref{app:training-appendix} provides details for reproducibility.\footnote{Our code and pretrained models are available at \url{https://github.com/nweir127/COD3S}}
\mysubsection{Evaluation}
We show that \method{} induces sensible inference of diverse but relevant semantic bins and causal statements. Examples of generation are shown in \autoref{app:ranked_gen} and additionally \autoref{app:bin-examples}.
\begin{table}[!t]
    \centering
    \small
    \setlength\tabcolsep{4pt}
    \begin{tabular}{llcc}
\toprule
    \multicolumn{2}{c}{\multirow{1}*{\textbf{COPA}}} & \multicolumn{1}{c}{\textbf{C $\rightarrow$ E}} &  \multicolumn{1}{c}{\textbf{E $\rightarrow$ C}} \\
    \multicolumn{2}{c}{\multirow{1}*{\textbf{3-Sets}}} & \textbf{BL-1} / \textbf{BL-2} / \textbf{SB} & \textbf{BL-1} / \textbf{BL-2} / \textbf{SB} \\
    \midrule
    \midrule
    \multicolumn{4}{l}{\textit{Baselines}}  \\
    \multicolumn{2}{l}{S2S} & 50.9 / 61.2 / .397 & 58.1 / 71.4 / .464 \\
\multicolumn{2}{l}{S2S + Sigs} & 46.7 / 58.5 / .323 & 50.7 / 65.3 / .326 \\
\midrule
\multicolumn{4}{l}{\textit{Other Decoding Methods}}  \\ 
\multicolumn{2}{l}{DPC (Li et al.)} & 49.2 / 58.1 / .389 & 57.4 / 67.0 / .425 \\
\multicolumn{2}{l}{S2S-RS (Li et al.)} & 78.2 / 90.3 / .635 & 75.4 / 89.7 / \textbf{.632} \\
\multicolumn{2}{l}{S2S-RS} & \textbf{83.6} / 95.7 / \textbf{.735} & \textbf{78.5} / \textbf{91.3} / \textbf{.639} \\
\midrule
\multicolumn{4}{l}{\textit{Two-Step \method{} Inferences}} \\
    \textbf{Sig} & \textbf{Sent} & & \\
    Beam & Beam & 79.1 / 93.2 / .618 & 70.6 / 84.8 / \textbf{.625} \\
Beam & MMI & 77.0 / 91.9 / .634 & 72.2 / 85.0 / \textbf{.613} \\
MMI & MMI & 73.6 / 87.9 / .608 & 72.0 / 85.3 / .586 \\
MMI & MMI-RS & \textbf{84.2} / \textbf{97.1} / .657 & \textbf{76.6} / \textbf{89.4} / \textbf{.617} \\
\multicolumn{2}{r}{\textit{$-$ Ham Heur}} & \textbf{81.1} / 93.9 / .620 & 70.4 / 84.2 / .508 \\
\midrule
    \midrule 
    \end{tabular}
    \setlength\tabcolsep{5pt}
    \begin{tabular}{lcccccc}
     \multicolumn{1}{r}{\textbf{Cos Threshold}:} & \textbf{0 } & \textbf{.1} & \textbf{.25} & \textbf{.5} & \textbf{.75}  \\ \midrule
      S2S & 10.0 &  6.40 &  4.52 &  2.85 &  1.70 \\
      S2S + RS       & 10.0 &  9.99 &  9.86 &  7.93 &  3.47 \\
      \method{} +MMI +RS  & 10.0 &  9.89 &  9.44 &  6.55 &  2.54 \\
      \bottomrule
    \end{tabular}
    \caption{\textbf{(Upper)} Diversity metrics (\textbf{BLEU-1} / \textbf{BLEU-2} / \textbf{SBERT}) over 3-best decoded outputs.\textbf{ (Lower)} Count of semantically distinct effect outputs out of 10, with  duplicates ruled out using SBERT cosine.}
    \label{tab:copa-diversity-scores}
\end{table}
We quantitatively compare \method{} against the outputs of regular seq2seq beam search, as well as of lexically constrained decoding with disjunctive positive constraints (DPC) and random sample decoding (S2S-RS) provided by \citeauthor{causalbank}\footnote{We also compare against our own S2S-RS using the same \proc{fairseq} model as the \method{} methods.} We included in the comparison instances of \method{} with and without MMI reranking, as well as with random sampling in place of beam search.

\paragraph{Automatic Diversity Metrics}
We use the formula of \citet{shu-etal-2019-generating}, which takes the pairwise average of dissimilarity score $\Delta$ over output set $Y$.
\vspace{-0.2em}
\[
\text{Diversity}(Y) = \dfrac{1}{|Y|(|Y| -1 )} \sum_{\substack{y,y' \in Y; \
          y\neq y'}} \Delta(y, y')
\vspace{-0.3em}
      \] 
To measure \textit{lexical} diversity, we set $\Delta(y,y')$ to be the sentences' inverse ($100$ minus) BLEU-1 and -2 scores.\footnote{Implemented using the SacreBLEU toolkit~\cite{post-2018-call}.} To measure \textit{semantic} diversity, we set $\Delta$ to be the cosine distance between their SBERT embeddings. Higher scores imply greater diversity. Following \citeauthor{causalbank}, we evaluate on 100 examples from an out-of-distribution dev split of the Choice of Plausible Alternatives dataset~\cite[COPA;][]{gordon-etal-2012-semeval}, with results shown in \autoref{tab:copa-diversity-scores}.\footnote{Results over 10 outputs and over a within-distribution train split from CausalBank are shown in Appendix \autoref{tab:cb-diversity-scores}.} In both cases, \method{}  outperforms all other methods except random sampling, the addition of which also improves the diversity of \method{} itself.\footnote{We verified the significance of numerical results using Wilcoxon two-sided signed-rank tests implemented \href{https://docs.scipy.org/doc/scipy/reference/generated/scipy.stats.wilcoxon.html}{via SciPy} with p=.05.} 
We also use the SBERT diversity score to \textit{count} semantically diverse outputs by marking as duplicates those for which the embedding of the completed phrase (``X $\dots$ Y") falls below some distance threshold from that of an earlier candidate. \autoref{tab:copa-diversity-scores} (lower) shows that both the best \method{} model as well as random sampling produce far more semantically distinct statements than the beam search baseline.
\begin{figure}
\includegraphics[width=\columnwidth]{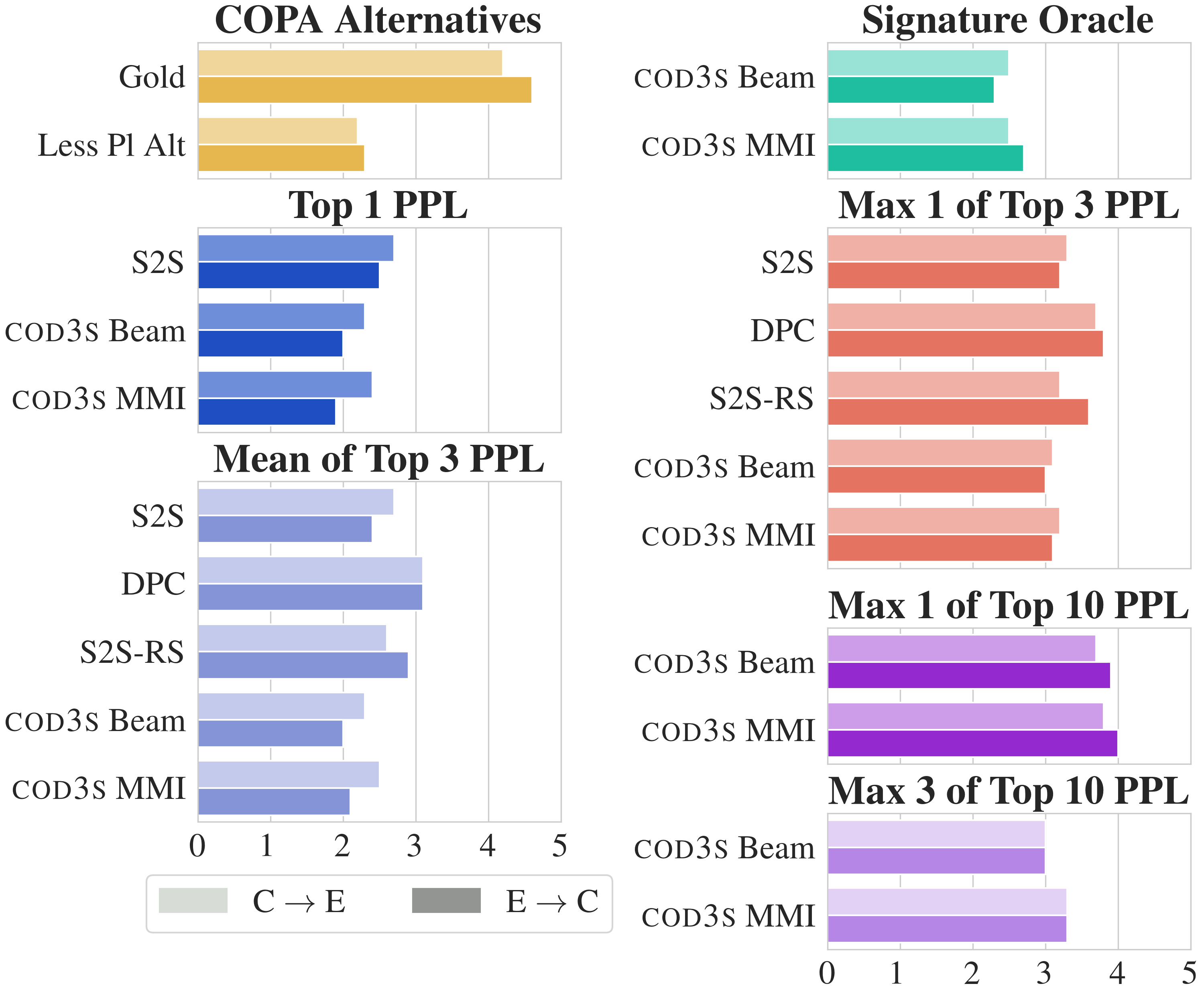}
\caption{
Results of human evaluation of plausibility. Ratings are shown in comparison to the gold answer and less plausible alternative from COPA. Mean/max ratings per input are presented for $1,3$-best outputs ranked by forward score (PPL).
To demonstrate that \method{} produces plausible response from many semantic bins, we also show max ratings from top-10 outputs. }
\label{fig:human-eval}
\end{figure}
\begin{table*}[t]
    \centering
    \small
    \begin{tabular}{lll}
    \toprule
         \multicolumn{3}{l}{\textbf{Cause Input: my favorite song came on the radio}}  \\ \midrule 
         Bin Medoid & \textit{I will try this version for sure} & \textit{I was quite excited to finally experience it} \\ \cmidrule(lr){2-2} \cmidrule(lr){3-3}
        Ranked Predictions &  \textbf{I decided to listen to it} & \textbf{I was excited to hear it again} \\
         & I decided to hear it & I was pleasantly surprised to hear it \\
         & I figured I'd try it & I'm glad to see it here\\
         \midrule
         \multicolumn{3}{l}{\textbf{Effect Input: the executive decided not to hire the applicant}} \\ \midrule 
         Bin Medoid & \textit{I knew that they expected it} & \textit{they are what earn you cash} \\ \cmidrule(lr){2-2} \cmidrule(lr){3-3}
       Ranked Predictions  &  \textbf{they knew she was not qualified} & \textbf{they could not afford the payments} \\
         & they knew it would be a mistake & it would cost them money \\
         & she knew she had to & she was paid \\
         \bottomrule
    \end{tabular}
    \caption{Examples of generation conditioned on semantic bins. Predictions are ranked according to maximum mutual information (MMI) and shown aside the given bin's representative medoid. }
    \label{app:ranked_gen}
\end{table*}
\paragraph{Human Evaluation}
Our automatic metrics 
quantify diversity without tracking task effectiveness, which we evaluate by collecing judgments on Amazon Mechanical Turk. 
We ask workers to judge the plausibility of responses as causal completions (on a 0-5 Likert scale). For all methods except \method{}, we use the exact outputs evaluated in \citet{causalbank} and provided to us by the authors. The response sets for these models contain the top 3 decoded sentences under perplexity (PPL).
We compare these to the top 3 as well as the top 10 sentences decoded by \method{} with and without MMI re-ranking (signature and sentence, no random sampling) ordered by PPL of the signature tokens. This discrepancy in per-model outputs reflects that we seek to evaluate \method{}, which is specifically crafted to produce a large set of distinct viable candidates, as directly as possible against the \citet{causalbank} responses from models that are not necessarily crafted with the identical aim. Naturally occurring propositions have far more than 10 plausible and distinct causes and effects, and so we would hope that the $10^{\text{th}}$ output of our one-to-many model would have similar quality to the $1^\text{st}$ of the other models.

Results are shown in \autoref{fig:human-eval}.\footnote{A tabular form of the results is given in Appendix \autoref{tab:human-eval}.} We observe that top 1 and 3 \method{} responses according to PPL (blue) are comparable albeit slightly lower on average than those of the other models.\footnote{DPC and S2S-RS output PPLs were not provided by \citeauthor{causalbank}, so they are omitted from top-1 comparison.} This may partially be attributed to the difficulty of the signature inference step, in which the differences in the top 100 predicted binary sequence PPLs are typically small. A \method{} `oracle' that conditions generation on the gold answer's signature (which often has low predicted likelihood) performs more competitively (green). 

We find that at least 1 of the top 3 signatures predicted by \method{} yields a competitively plausible sentence; when we take the highest plausibility score from the top 3 of each model under their respective PPL orderings (red), \method{} and baseline S2S to be interchangeable. 
If we expand to the larger set of 10 outputs for \method{} models, we find that the mean of the 3 highest plausibility scores (faded purple) for the MMI model is comparable to the 1 best of the base seq2seq (red) and better than the mean of the top 3 PPL (faded blue) for any model. This indicates that the 10 output set, which shows under automatic metrics to contain higher numbers of semantically diverse statements, also contains at worst a set of 3 outputs that are better than the 3 from models not designed for one-to-many diverse prediction.


\paragraph{Qualitative Analysis}
\autoref{app:ranked_gen} shows examples of models predicting and re-ranking sentences within inferred signature bins. Candidate predictions listed in order of MMI score reflect the ability of MMI-based reranking to select the candidates within a bin that are most relevant to the input. Outputs are shown beneath a representative bin \textit{medoid}, i.e. the sentence with minimized embedding cosine distance from all other training sentences that fall in the bin. 
The two-step inference process depicted here allows for a level of interpretability on the signature level, as sampling training sentences from the inferred semantic bin gives a snapshot of an inferred semantic space that can be more informative than individual sentences alone.
 
Future work might explore alternative methods for signature inference.  The bit sequence likelihoods predicted by \method{} are often clumped together
and/or biased towards signatures that intuitively do not apply to an input but are over-represented in the training set.
We also observe that although MMI decoding discourages bland context insensitive statements, there is still a model tendency towards a small set of generic predicates, e.g. `having,' `knowing,' or `being able to.' 
 
\section{Conclusion}
\label{sec:conclusion}
We have outlined \method{}, a method for producing semantically diverse statements in open-ended generation tasks. We design sentence LSH signatures that encode bitwise the semantic similarity of underlying statements; conditioning generation on different signatures yields outputs that are semantically heterogeneous. \method{} leads to more diverse outputs in a multi-target generation task in a controllable and interpretable manner, suggesting the potential of semantically guided diverse decoding for a variety of text generation tasks in the future.


\section{Acknowledgments}
We thank the reviewers for their insightful comments. We additionally thank Elizabeth Salesky, Huda Khayrallah, Rachel Wicks and Patrick Xia for their discussions, feedback and compute. 
This work was supported in part by DARPA KAIROS
(FA8750-19-2-0034). The views and conclusions contained in this work are those of the authors and should not be interpreted as representing official policies or endorsements by DARPA or the U.S. Government.

\bibliographystyle{acl_natbib}
\bibliography{emnlp2020,anthology}
\newpage

\appendix
\section{Random Hyperplane LSH Details}
\label{app:LSH}
\begin{figure}[h!]
    \centering
    \includegraphics[width=.47\textwidth]{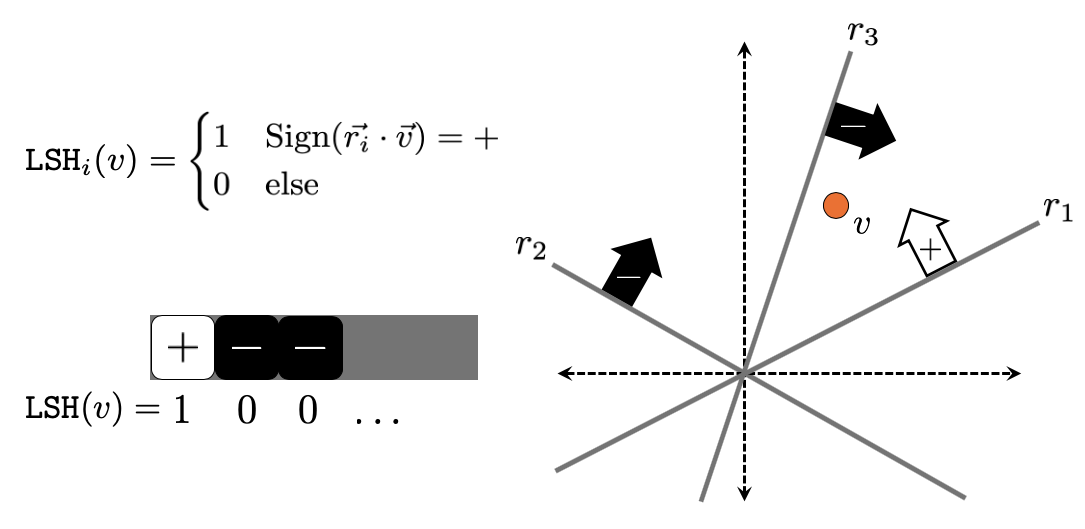}
    \caption{Computation of a 2D vector $v$'s LSH bit signature as the signs of the dot products with $d$ random normal vectors $r_{1},\dots,r_{b}$. }
    \label{fig:lsh}
\end{figure}

The popular LSH variant introduced by \citet{charikar2002similarity} leverages \textit{random hyperplane projections} to compute discrete $b$-length bit signatures.  Each individual bit is determined from the sign of the dot product between a given embedding and one of a set of $b$ pre-computed random normal vectors. 
One geometric intuition is that the hyperplane implied by each random normal vector partitions the full embedding space in half, and the sign of the dot product designates the partition into which the input embedding falls. This is illustrated in \autoref{fig:lsh} using a simplified case with a 2-D vector $v$ and three random vectors $r_1,r_2,r_3$ indicating partitions of the Cartesian plane.\footnote{Figure adapted from \href{http://cs.jhu.edu/~vandurme/papers/VanDurmeLallACL10-slides.pdf}{slides} of \citet{van-durme-lall-2010-online} with permission of the authors.}

Formally, given a set of high-dimensional vectors in $\mathbbm{R}^D$, we randomly sample $b\ll D$ random vectors $\vec{r}_1, \dots \vec{r}_d$ from the $D-$dimensional Gaussian distribution.
Then, given a high-dimensional embedding $\vec{v}$, we construct the $b$-bit signature $\texttt{LSH}(v) = [\texttt{LSH}_1(v), \dots \texttt{LSH}_d(v)]$ using the hash functions
$$
    \texttt{LSH}_{i}(v) = \begin{cases} 1 & \text{if } \vec{r_i} \cdot \vec{v} \geq 0 \\
0 & \text{if } \vec{r_i} \cdot \vec{v} < 0 \end{cases}
$$

The number of matching bits in the signatures of two vectors $u,v$ provides an estimate of their \textit{hash collision probability}, i.e. the likelihood that they fall in the same partition of any random hyperplane. 
This probability is provably\footnote{\citet{charikar2002similarity,li2013coding}} monotonically increasing with the vectors' inner product.  
\citet{goemans-1995-improved} similarly prove that the Hamming distance between signatures is proportional to the angle between the vectors, which correlates highly with cosine distance barring high discrepancies in vector norms.

\section{Training Details}
\label{app:training-appendix}
\begin{verbatim}
fairseq-train 
 --adam-betas "(0.9, 0.98)"
 --arch transformer_iwslt_de_en 
 --criterion 
   label_smoothed_cross_entropy
 --label-smoothing 0.1 
 --dropout 0.1 --weight-decay 0 
 --bpe sentencepiece 
 --optimizer adam  --clip-norm 0.1 
 --lr 5e-4 --lr-scheduler inverse_sqrt 
 --warmup-updates 4000 
 --max-epoch 10 
 --share-all-embeddings 
\end{verbatim}
We train models with \proc{fairseq} using the  \texttt{transformer\_iwslt\_de\_en} architecture. We use 6 encoder and decoder layers with 512-dimensional hidden states and shared embedding layers (a total of 36.6M trainable parameters). Signature tokens are assigned special tokens during BPE encoding.
We train models for 10 epochs with an early stopping patience of 2 validations. We use the Adam optimizer~\cite{kingma2014adam} with $0.1$-smoothed cross entropy loss, a $\num{5e-4}$ learning rate with inverser square root scheduling, $0.1$ dropout and $0.1$ norm clipping. All other training parameters were the \proc{fairseq} defaults at the time of submission.
We observe performance drops when 1) norm clipping threshold is not sufficiently low, 2) BPE vocabulary size is 32K instead of 10K, and 3) weight decay is set to .001. Training takes roughly 12 hours on two Titan 24GB RTX GPUs for each of four models (two forward, two backward for MMI reranking).

Backwards scoring models for MMI-bidi are trained with the opposite dataset as their corresponding forward models;  we find training most effective when the data's syntactic direction (``X \dots Y'') matches the direction of inference
(X $\rightarrow$ Y). In other words, all C$\rightarrow$ E models are trained on ``X, so Y'' data regardless of their use as forward or backward scoring models.  We used the ``X because Y'' training split from \citet{causalbank}. We constructed the 10M ``X so Y'' examples ourselves: we took a 20M random sample of all such examples in the dataset, filtered to remove sentences a) containing numerical and special characters or b) containing either a source or target with greater than 12 tokens, and then downsampled the remaining set to a 10M/4K/4K train/dev/test split.  
\begin{table}[!t]
    \setlength\tabcolsep{4pt}
    \centering
    \small
    \begin{tabular}{llcc}
    \toprule
    \multicolumn{2}{c}{\multirow{1}*{\textbf{Causalbank}}} & \multicolumn{1}{c}{\textbf{C $\rightarrow$ E}} &  \multicolumn{1}{c}{\textbf{E $\rightarrow$ C}} \\
    \multicolumn{2}{c}{\multirow{1}*{\textbf{3-Sets}}} & \textbf{BL-1} / \textbf{BL-2} / \textbf{SB} & \textbf{BL-1} / \textbf{BL-2} / \textbf{SB} \\
    \midrule
    \midrule
    \multicolumn{4}{l}{\textit{Baselines}}  \\
    \multicolumn{2}{l}{S2S} & 54.2 / 62.9 / .348 & 59.8 / 71.4 / .428 \\
\multicolumn{2}{l}{S2S + Sigs} & 47.5 / 56.6 / .248 & 56.2 / 70.3 / .302 \\
\midrule
\multicolumn{4}{l}{\textit{Other Decoding Methods}}  \\ 
\multicolumn{2}{l}{DPC (Li et al.)} & 41.8 / 49.4 / .293 & 47.4 / 55.3 / .319 \\
\multicolumn{2}{l}{S2S-RS (Li et al.)} & 77.5 / 89.3 / .567 & \textbf{82.6} / \textbf{94.1} / .622 \\
\multicolumn{2}{l}{S2S-RS} & \textbf{87.0} / \textbf{96.8} / \textbf{.676} & 82.1 / 92.1 / \textbf{.626} \\
\midrule
\multicolumn{4}{l}{\textit{Two-Step \method{} Inferences}} \\
    \textbf{Sig} & \textbf{Sent} & & \\
    Beam & Beam & 84.0 / 94.2 / .603 & 77.1 / 89.6 / .558 \\
Beam & MMI & 80.0 / 90.9 / .571 & 74.0 / 86.3 / .542 \\
MMI & MMI & 75.1 / 86.6 / .554 & 70.7 / 83.9 / .543 \\
MMI & MMI-RS & \textbf{85.9} / \textbf{95.4} / .620 & 78.1 / 90.9 / .563 \\
\multicolumn{2}{r}{\textit{$-$ Ham Heur}} & 80.4 / 90.8 / .521 & 74.0 / 87.8 / .501 \\
\bottomrule
    \toprule
    \multicolumn{2}{c}{\multirow{1}*{\textbf{COPA}}} & \multicolumn{1}{c}{\textbf{C $\rightarrow$ E}} &  \multicolumn{1}{c}{\textbf{E $\rightarrow$ C}} \\
    \multicolumn{2}{c}{\multirow{1}*{\textbf{10-Sets}}} & \textbf{BL-1} / \textbf{BL-2} / \textbf{SB} & \textbf{BL-1} / \textbf{BL-2} / \textbf{SB} \\
    \midrule
    \midrule
    \multicolumn{4}{l}{\textit{Baselines}}  \\
    \multicolumn{2}{l}{S2S} & 59.9 / 71.5 / .466 & 62.5 / 76.7 / .509 \\
\multicolumn{2}{l}{S2S + Sigs} & 52.4 / 64.8 / .360 & 55.3 / 70.0 / .397 \\
\multicolumn{2}{l}{S2S-RS} & \textbf{84.7} / \textbf{96.9} / \textbf{.746} & \textbf{83.8} / \textbf{95.1} / \textbf{.693} \\
\midrule
\multicolumn{4}{l}{\textit{Two-Step \method{} Inferences}} \\
    \textbf{Sig} & \textbf{Sent} & & \\
    Beam & Beam & 81.7 / 95.5 / .658 & 75.8 / 89.6 / .660 \\
Beam & MMI & 78.5 / 93.1 / .653 & 75.1 / 89.2 / .639 \\
MMI & MMI & 75.8 / 90.6 / .633 & 74.3 / 88.2 / .612 \\
MMI & MMI-RS & 82.6 / 96.1 / .676 & 78.2 / 91.8 / .647 \\
\multicolumn{2}{r}{\textit{$-$ Ham Heur}} & 80.5 / 93.8 / .619 & 72.5 / 86.2 / .544 \\
\bottomrule
    \end{tabular}
    \caption{Automatic diversity metrics (\textbf{1-BLEU} / \textbf{2-BLEU} / \textbf{SBERT}) evaluated over the outputs of 16-bit \method{} and other decoding methods. Results are shown for 3-best outputs over 100 in-distribution CausalBank examples and 10-best over out-of-distribution COPA. Following \citet{causalbank}, the same 100 ``X because Y'' pairs were used to evaluate models of both inference directions. }
    \label{tab:cb-diversity-scores}
    \end{table}
    
\section{Decoding According to Semantic Bins}
\label{app:bin-examples}
We experimented with bit lengths of 8, 16, and 32, and found the middle value to best balance specificity with accuracy. We also explored a variant that merged signatures into a single token rather than treating them as token-per-bit, but found the model to perform qualitatively worse. We experimented with Hamming distance heuristic thresholds of 0 through 6 and found the best value (2) for 16-bit \method{} using qualitative analysis of side-by-side predictions. The MMI-bidi $\lambda_s,\lambda_y$ values were found using simple grid search, comparison of automatic metrics, and side-by-side analysis. The nature of the output set is sensitive to only large changes (orders of magnitude) in $\lambda_s$ values, as the likelihoods of signature sequences are rather close in value; however, smaller, 0.1-increment changes to the sentence weight $\lambda_y$ showed to have a greater effect on the relevance and specificity of output causes/effects. This comports with results from previous applications of MMI-bidi decoding for sentences~\cite{li-etal-2016-diversity}. 

\autoref{app:full-outputs} shows side-by-side outputs of models with and without MMI re-ranking conditioned on the same n-best inferred signatures. \autoref{tab:cb-diversity-scores} shows results of automatic diversity evaluation on the in-distribution training sample from CausalBank following \citet{causalbank}.  \autoref{tab:human-eval} provides a tabular version of the human plausibility scores depicted in \autoref{fig:human-eval}.

\begin{table}[!t]
    \setlength\tabcolsep{5pt}

    \small
    \centering
    \begin{tabular}{lcccc}
    \toprule
    \multicolumn{3}{l}{\textbf{C $\rightarrow$ E} / \textbf{E $\rightarrow$ C} 
    \hspace{.18cm} 
    \textbf{Gold:}  4.2 / 4.6} & \multicolumn{2}{c}{\textbf{Pl Alt:}  2.2 / 2.3}   \\ \midrule
    &  \multicolumn{2}{c}{\textbf{Top PPL}} &  \multicolumn{2}{c}{\textbf{Max Score}} \\ \cmidrule(lr){2-3}\cmidrule(lr){4-5}
    \multicolumn{1}{l}{{\textbf{Method}}} &\multicolumn{1}{c}{T1} & \multicolumn{1}{c}{T3} & \multicolumn{1}{c}{T1} & \multicolumn{1}{c}{T3 \textit{(/ 10)}  } \\
    \midrule
    \midrule
    \multicolumn{1}{l}{S2S}   &  2.7 / 2.5 & 2.7 / 2.4	 &  3.3 / 3.2 & \\ 
    \multicolumn{1}{l}{DPC}   &  \multicolumn{1}{c}{---}  & 3.1 / 3.1	 &  3.7 / 3.8 &   \\
    \multicolumn{1}{l}{S2S-RS} & \multicolumn{1}{c}{---} & 2.6 / 2.9	 &  3.2 / 3.6 &  \\ 
    \midrule 
    \multicolumn{2}{l}{\method{} }         &  &  &                  \\
    \cmidrule(lr){1-1}
    \multicolumn{1}{l}{Beam} &  2.3 / 2.0 & 2.3 / 2.0 &  3.1 / 3.0 &  \\
    \multicolumn{1}{l}{\textit{(Oracle)}} & \textit{2.5 / 2.3} & & &  \\ 
    \multicolumn{1}{l}{\textit{(10 Outputs)}} &  & &  3.7 / 3.9 & 3.0 / 3.0 \\ 
    MMI   &  2.4 / 1.9 &  2.5 / 2.1 &  3.2 / 3.1 &  \\
     \textit{(Oracle)} &  \textit{2.5 / 2.7} & & \\
      \multicolumn{2}{l}{\textit{(10 Outputs)}}  & &  3.8 / 4.0 & 3.3 / 3.3 \\

    \bottomrule
    \end{tabular}
    \caption{Tabular form of human evaluation results displayed in \autoref{fig:human-eval}.}
    \label{tab:human-eval}
\end{table}
\begin{table*}[t!]
    \centering
    \small
    \begin{tabular}{llllll}
    \multicolumn{3}{l}{\textbf{Cause: the tenant misplaced his keys to his apartment}} &  \multicolumn{3}{l}{ \textbf{Effect: the man threw out the bread}} \\ \midrule
1 &         he couldn't leave the house &                    & 1 &              he didn't want to eat it &          \\
2 &    \textit{he couldn't get out of the house} &             \textit{Dupl. of 1 (.01)} & 2 &    he didn't like it &                   \\
3 &             he had to get a new one &                    & 3 &   \textit{ he didn't like the taste} &  \textit{Dupl. of 2 (.05)}  \\
4 &      he had to go back to the hotel &                    & 4 &    it was too much for him to handle &    \\
5 &            \textit{he had to find a new one} &             \textit{Dupl. of 3 (.02)} & 5 &  \textit{he didn't want to cook it} &       \textit{Dupl. of 1 (.07)}\\
6 &      \textit{he couldn't get into the house} &              \textit{Dupl. of 1 (.06)} & 6 & he didn't know how to cook it &       \\
7 &      {he had to go back to the house} &                    & 7 &   \textit{it wasn't good for him} &       \textit{Dupl. of 1 (.07)}\\
8 &      \textit{he couldn't leave the building} &              \textit{Dupl. of 1 (.02)} & 8 &            \textit{he didn't like how it tasted} &       \textit{Dupl. of 2 (.05)}\\
9 &  he had to go to the police station &                    & 9 &   \textit{he couldn't eat it} & \textit{Dupl. of 1 (.06)} \\
10 &  \textit{he had to go back to his apartment} &              \textit{Dupl. of 7 (.07)} & 10 &   it was overcooked &    \\ 
\bottomrule
    \end{tabular}
    \caption{Detection of duplicate causes and effects using a threshold SBERT embedding cosine distance of $0.1$. We embed the full ``X \dots Y'' statements so as to provide context to the paraphrase detection. Model outputs are those of a regular seq2seq. }
    \label{tab:detecting_paraphrases}
\end{table*}

\section{Counting Semantically Distinct Outputs using SBERT}
\label{app:threshold}
We construct a method for automatically counting the number of semantically diverse sentences in a candidate cause/effect set. We encode each prediction with the context of the input by taking the SBERT embedding of the completed sentence "X \{because, so\} Y." We then rule out all sentences whose embedding cosine distance from that of a higher-ranked candidate is lower than some threshold. We use a simple grid search over various threshold values 
and find that a value of $.1$ yields a sensitivity to paraphrastic cause/effect predictions similar to that of a human reader. As other tasks might merit different such thresholds, we provide multiple such counts in \autoref{tab:copa-diversity-scores}. \autoref{tab:detecting_paraphrases} shows example cases of duplicate detection among generated candidate sets.
\section{Cosine/LSH Hamming Correlations with STS and Bin Statistics}
\label{app:bins-appendix}
\autoref{tab:sts-analysis} shows the Spearman $\rho$ coefficient with STSbenchmark judgments for cosine and approximate LSH Hamming distances of embeddings for 
BERT, SBERT (and larger variant SRoBERTa), and pBERT~\cite{hu2019large}, a BERT model fine-tuned to predict paraphrastic similarity, albiet not via angular similarity of embeddings.
\autoref{tab:cluster-analysis} provides details regarding the distributions of sentences into LSH bins of differing levels of granularity using SRoBERTa-L embeddings. 
\section{Human Evaluation of Plausibility}
\label{app:human-eval}
We showed 200 COPA input statements (100 each for cause-to-effect and effect-to-cause) to Amazon Mechanical Turk workers and asked them to judge the plausibility of model predictions, specifically as completions of 
a causal statement
of the form
``X because Y'' or ``Y, so X.'' The order of the examples were randomized. Four annotators rated each input/prediction pair. We required annotators to have at least a 97\% approval rating, be located in the US, and have completed at least 500 HITs. Annotators were given an hour to complete each HIT. The median completion time for the task was 5 minutes, and workers were paid \$0.50 per HIT. We included at least two attention checks. 

\begin{table*}[]
    \centering
    \small
    \setlength\tabcolsep{4pt}
    \begin{tabular}{l l l }
    \toprule
    \textbf{W/O MMI Reranking}     & \textbf{ W/ MMI Reranking} & \textbf{\textit{Conditioned Bin Medoid}}  \\
    \midrule \midrule
    \multicolumn{3}{l}{\textbf{Cause: {I was confused by the professor's lecture}}}\\ 
    \multicolumn{3}{l}{\textbf{Gold Effect: {I asked the professor questions}}}\\  \midrule
         \textbf{I asked him about it}   &           \textbf{I asked a few questions } & \textbf{\textit{I need some feedback from you} } \textbf{(Gold bin)} \\
         I decided to try it    & I decided to look it up     &               \textit{I will try this version} \\
         I thought I'd ask here   &       I decided to ask the teacher      &   \textit{I might change them at some point} \\
         I decided to open it up  &                 I opened it up and started reading     &   \textit{you can check it out} \\ 
        I did my own research   &      I did a quick math lesson         &         \textit{it is easy to get everything aligned} \\
        \midrule
    \multicolumn{3}{l}{\textbf{Cause: {several witnesses of the crime testified against the suspect}}}\\
    \multicolumn{3}{l}{\textbf{Gold Effect: {the suspect was convicted}}}\\ \midrule
        \textbf{he's got that going for him }    &    \textbf{the case was taken to court} & \textbf{\textit{we did it this way}} \textbf{(Gold bin)} \\
         he knew what to do      &     the case was resolved    &   \textit{this is a simple solution that makes sense} \\
         the jury is still out & the jury was left to investigate  &  \textit{ everyone will know what it is } \\
         they didn't have to deal with it   &  there was no need for an attorney   &  \textit{I guess I won't have to think about this} \\
         it was easy to follow  &   the police proceeded to investigate  &    \textit{this recipe is ready to go } \\
         \midrule
    \multicolumn{3}{l}{\textbf{Cause: {the papers were disorganized}}}\\
    \multicolumn{3}{l}{\textbf{Gold Effect: {I put them into alphabetical order}}}\\ \midrule
        \textbf{I had to enter them }    &    \textbf{I had to print them out} & \textbf{\textit{the opening sequence was there}} \textbf{(Gold bin)} \\
         that's out of the question     &     I gave up on it   &                  \textit{I won't use it in anything anymore} \\
         I decided to skip it  &    I decided not to publish them       & \textit{I opted not to do any} \\
         I got a new one      &       I had to edit them   &               \textit{we came at a good time} \\
         we had to start all over again  &    I had to start all over again    &  \textit{it should be open by then} \\
    \midrule \midrule
         \multicolumn{3}{l}{\textbf{Effect: {the woman hired a lawyer}}}\\
    \multicolumn{3}{l}{\textbf{Gold Cause: {she decided to sue her employer}}}\\ \midrule
        \textbf{she wanted to}    &    \textbf{she wanted a lawyer} & \textbf{\textit{they want to crack down on it}} \textbf{(Gold bin)} \\
         she thought she could win      &     she wanted to be in charge of her case    &   \textit{it can be an ideal method for you to succeed} \\
         she had a plan   &    she felt she had enough evidence  &    \textit{it was what we had and it turned out fine} \\
        she trusted him   &  she wanted to help people   &  \textit{I did trust and respect the person} \\
         she wanted to be a mother    &           she wanted to protect her family      & \textit{all ages enjoy them}      \\
         \midrule
         \multicolumn{3}{l}{\textbf{Effect: {I avoided giving a straight answer to the question }}}\\
    \multicolumn{3}{l}{\textbf{Gold Cause: {the question made me uncomfortable}}}\\ \midrule
        \textbf{I didn't want to offend anyone}   &    \textbf{I didn't want to offend anyone} & \textbf{\textit{I didn't like to speak}} \textbf{(Gold bin)} \\
         I didn't understand it      &     I didnt know what I was talking about    &   \textit{I didn't understand them} \\
         there was no one to talk to & I didn't want to talk about it   &   \textit{I'm not allowed to talk to them about anything} \\
       the answer was obvious  & I thought the answer would be obvious  &             \textit{everyone's familiar with it} \\
         I was so embarrassed  & I thought I was stupid    &  \textit{it looked ridiculously saturated}   \\
         \midrule
         \multicolumn{3}{l}{\textbf{Effect: {I learned how to play the board game}}}\\
    \multicolumn{3}{l}{\textbf{Gold Cause: {my friend explained the rules to me}}}\\ \midrule
        \textbf{I learned a lot about the game}   &    \textbf{I wanted to learn to play the game} & \textbf{\textit{it offers some good information}} \textbf{(Gold bin)} \\
         i felt like it   &  i felt i had to  &  \textit{I feel it to be so} \\
         it was so easy &  it was easy to play  &  \textit{it is done nicely and realistically} \\
       it worked &  i knew i was going to play it   &  \textit{they have now got it right} \\
         I love to play online  &  I wanted to play online  &   \textit{the online wants anyone spreading the phrase } \\
         \bottomrule
    \end{tabular}
    \caption{Example \method{} output responses with and without MMI-bidi sentence re-ranking. Predictions are shown alongside their conditioned bin's representative medoid sentence. ``Bin oracle'' predictions conditioned on the signature of gold sequence (\textbf{Gold bin}) are shown for comparison. }
    \label{app:full-outputs}
\end{table*}
\begin{table*}[!t]
    \centering
\small
\begin{tabular}{l|rrrrrrrr}
\toprule
bits &    4 &    8 &   16 &   32 &   64 &  128 &  256 & full \\
\midrule
BERT-B           & 0.01 & 0.08 & 0.11 & 0.12 & 0.09 & 0.14 & 0.15 & 0.13 \\
pBERT-B          & 0.05 & 0.09 & 0.09 & 0.11 & 0.13 & 0.14 & 0.15 & 0.14 \\
SBERT-B     & 0.41 & 0.51 & 0.61 & 0.69 & 0.76 & 0.80 & 0.82 & 0.85 \\
SBERT-L    & 0.42 & 0.51 & 0.64 & 0.72 & 0.77 & 0.80 & 0.82 & 0.85 \\
SRoBERTa-B  & 0.38 & 0.51 & 0.61 & 0.71 & 0.77 & 0.81 & 0.83 & 0.85 \\
SRoBERTa-L & 0.42 & 0.55 & 0.65 & 0.74 & 0.80 & 0.83 & 0.85 & 0.86 \\
\bottomrule
\end{tabular}
\caption{Spearman $\rho$ correlation of LSH Hamming-based cosine approximations with human STS judgements on STSBenchmark (as well as cosine similarity of the full 768/1024-dimension embeddings)}
\label{tab:sts-analysis}
\end{table*}
\begin{table*}
\small
\begin{tabular}{r|rrrrrrrr}
\toprule
LSH Bits &                                4  &                                 8  &                                 12 &                                16 &                               20 &                               24 &                              28 &                              32 \\
\midrule
\multirow{2}*{\shortstack[r]{Distinct Sentences / \\ Populated Bin}} &       \multicolumn{1}{r}{ 5.55e5} &        \multicolumn{1}{r}{ 3.47e4} &       \multicolumn{1}{r}{ 2166.97} &       \multicolumn{1}{r}{ 135.85} &       \multicolumn{1}{r}{ 10.75} &        \multicolumn{1}{r}{ 2.47} &       \multicolumn{1}{r}{ 1.33} &       \multicolumn{1}{r}{ 1.10} \\
             &  \multicolumn{1}{r}{$\pm$ 1.91e5} &   \multicolumn{1}{r}{$\pm$ 2.37e4} &  \multicolumn{1}{r}{$\pm$ 2671.91} &  \multicolumn{1}{r}{$\pm$ 225.40} &  \multicolumn{1}{r}{$\pm$ 22.32} &   \multicolumn{1}{r}{$\pm$ 4.62} &  \multicolumn{1}{r}{$\pm$ 1.51} &  \multicolumn{1}{r}{$\pm$ 0.72} \\
\multirow{2}*{\shortstack[r]{Distinct Unigrams / \\ Populated Bin}} &        \multicolumn{1}{r}{1.28e5} &         \multicolumn{1}{r}{2.15e4} &        \multicolumn{1}{r}{3191.00} &        \multicolumn{1}{r}{415.27} &        \multicolumn{1}{r}{54.42} &        \multicolumn{1}{r}{15.71} &        \multicolumn{1}{r}{9.24} &        \multicolumn{1}{r}{7.87} \\
              &  \multicolumn{1}{r}{$\pm$ 2.24e4} &  \multicolumn{1}{r}{$\pm$ 8446.11} &  \multicolumn{1}{r}{$\pm$ 2378.42} &  \multicolumn{1}{r}{$\pm$ 430.38} &  \multicolumn{1}{r}{$\pm$ 73.41} &  \multicolumn{1}{r}{$\pm$ 19.10} &  \multicolumn{1}{r}{$\pm$ 6.63} &  \multicolumn{1}{r}{$\pm$ 3.64} \\
\% Buckets Populated             &                            100 &                             100 &                             100 &                             99.69 &                            78.73 &                            21.45 &                            2.49 &                            0.19 \\
STS $\rho$                       &                              0.42 &                               0.55 &                               0.61 &                              0.65 &                             0.69 &                             0.71 &                            0.73 &                            0.74 \\
\bottomrule
\end{tabular}

\caption{Analysis of bin clusters using the effects of 10 million CausalBank "X because Y" pairs.}
\label{tab:cluster-analysis}
\end{table*}
\begin{figure*}[t]
\includegraphics[width=\textwidth]{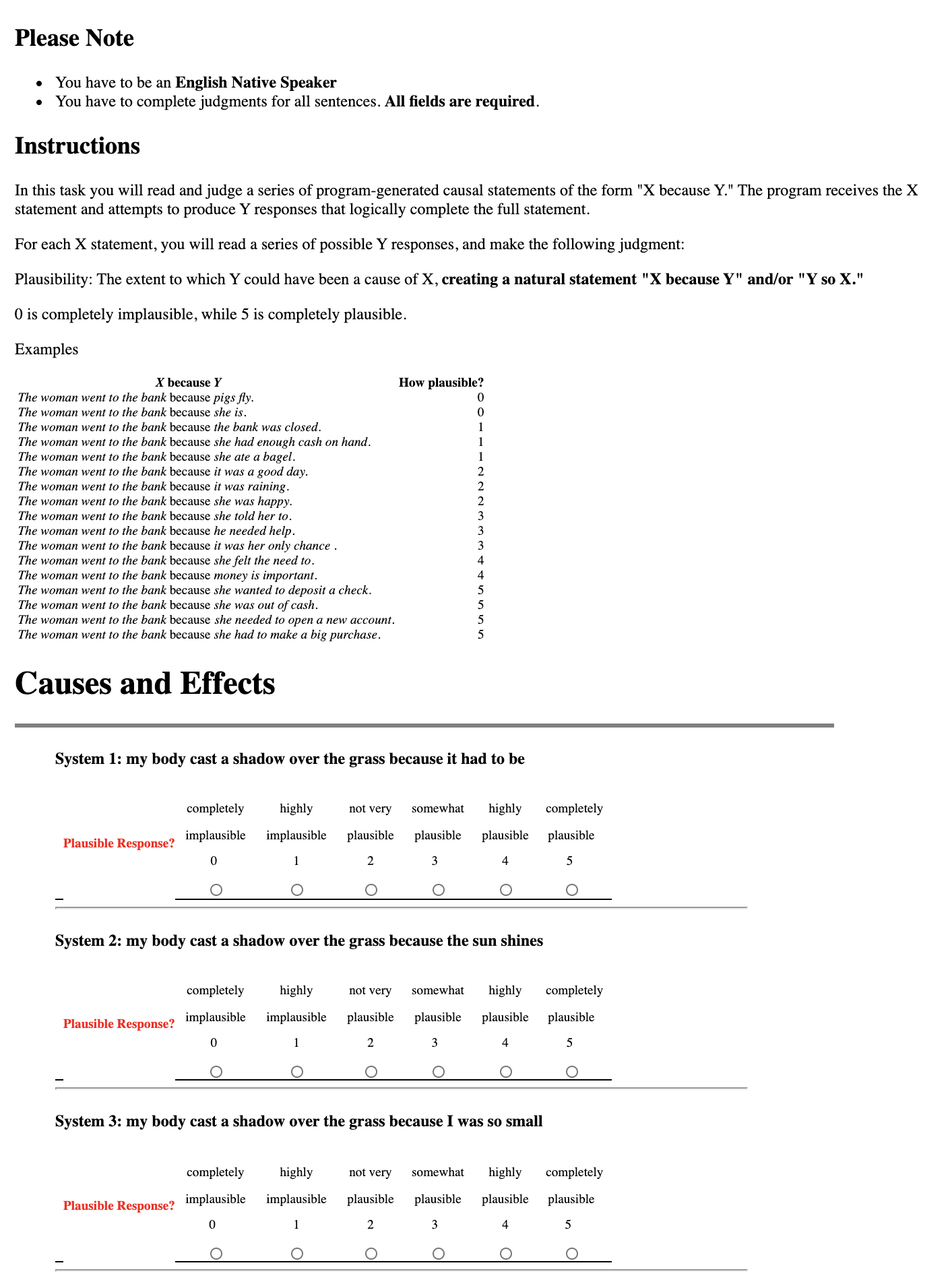}
\caption{Interface shown to Amazon Mechanical Turk workers during collection of plausibility judgments.}
\end{figure*}

\end{document}